\documentclass[a4paper,10.5pt]{elsarticle}
\usepackage{geometry}
\usepackage{makecell}
\usepackage{extarrows}
\usepackage{subfigure}
\usepackage{mathrsfs}
\usepackage{lineno}
\usepackage{ulem}
\usepackage{graphicx}
\usepackage{subfigure}
\usepackage{makecell}
\usepackage{fancyhdr}
\usepackage{amssymb}
\usepackage{multicol}
\usepackage{amsmath}
\usepackage[figuresright]{rotating}

\usepackage{hyperref}
\usepackage{cleveref}
\begin{document}
% \linenumbers
\begin{frontmatter}

\title{Clustering by the way of atomic fission}

\author{Shizhan Lu$^{1*}$}% , Longsheng Cheng$^1$, Haiyan Xu$^2$, Rashid Mehmood$^3$, Yu Han$^2$}

\address{ \small  $^1$College of Economics and Management, Nanjing University of Science and Technology, Nanjing,  210094,
 China (e-mail: lubolin2006@163.com) \\

%\small  $^2$College of Economics and Management, Nanjing University of Aeronautics and Astronautics, Nanjing 211106,  China \\

%\small  $^3$Department of Software Engineering, University of Kotli, AJ$\&$K 11100, Pakistan. \\
}

% \cortext[t1]{E-mail: Shizhan Lu (lubolin2006@163.com, lushizhan20140910@126.com)}

\begin{abstract}
Cluster analysis which focuses on the grouping and categorization of similar elements is widely used in various fields of research. Inspired by the phenomenon of atomic fission, a novel density-based clustering algorithm is proposed in this paper, called fission clustering (FC). It focuses on mining the dense families of a dataset and utilizes the information of the distance matrix to fissure  clustering dataset into subsets. When we face the dataset which has a few points surround the dense families of clusters, K-nearest neighbors local density indicator is applied to distinguish and remove the points of sparse areas so as to obtain a dense subset that is constituted by the dense families of clusters.  A number of frequently-used datasets were used to test the performance of this clustering approach, and to compare the results with those of algorithms. The proposed algorithm is found to outperform other algorithms in speed and accuracy.

\end{abstract}
\end{frontmatter}

\begin{multicols}{2}
\section{Introduction}

Cluster analysis is widely used in a number of different areas, such as climate research  \cite{PT}, computational biology, biophysics and bioinformatics \cite{EMB, HW}, economics and finance \cite{HJD, LG},  and neuroscience \cite{GS, AD}.
The basic task of clustering is to divide data into distinct groups on the basis of their similarity. Clustering methods can be categorized as density-based  \cite{BD, RA, MR}, grid-based \cite{PM}, model-based \cite{CT, MAK}, partitioning \cite{MA, LK}, and hierarchical \cite{MF, JD} approaches.

Initial methods of clustering tend to focus on finding the center point of every category, and then assigning the other points to the nearest center.  To make computer cluster data faster, some researchers, such as Schikuta \cite{SE}, Ma and Chow \cite{ME} et al, apply the grid-clustering method to divide objects part by part. The grid-clustering method does not need to cluster data point by point; however, this method is influenced by the size of grid cells and can not easily determine the number of categories.

Inspired by the phenomenon and rapid process of atomic fission, this paper proposed a fast and effective clustering algorithm, which we call fission clustering (FC). If the distances between every pair of  clusters are large enough, two maximal value are applied to determine the number of categories, ``the maximal crack of the distance matrix and the maximal value of all the distances between objects and their nearest neighbors''.  Otherwise, the K-nearest neighbors method  is applied to obtain a local density indicator for every object in the clustering dataset. Then the objects that have a small indicator value will be removed, and a dense subset with large distances between every two clusters is obtained.

\section{Related work}

Clustering is a classical issue in data mining. In recent decades, a number of typical clustering algorithms have
been proposed, such as DBSCAN \cite{EM}, OPTICS \cite{AM} (density-based);  STING \cite{WW}, CLIQUE \cite{AGR} (grid-based); Gaussian mixture models \cite{FRC}, COBWEB \cite{FDH} (model-based); K-means \cite{MJ},  CLARANS \cite{NRT} (partitioning) and DIANA \cite{KAL}, BIRCH \cite{ZHT} ( hierarchical).

Of the earlier methods found, the most representative clustering method is
K-means \cite{MJ}, which focuses on dividing data points into K
clusters by using Euclidean distance as the distance metric. K-means has many variants (see \cite{SR, TG}).  It also is applied as a useful tool for other method, for instance spectral clustering \cite{ZD}, the spectral clustering maps the data points
to a low-dimensional feature space to replace the Euclidean space in conventional K-means clustering. It can be reformulated as the weighted
kernel K-means clustering method.

More recently, an fast algorithm that finds density peaks (DP) was proposed \cite{RA} and widely used. It  combines the advantages of both density-based and centroid-based clustering methods. Many variants have since been developed by using DP, such as gravitation-based density peaks clustering algorithm \cite{JJH}, FKNN-DPC \cite{XJY} and SNN-DPC \cite{LIUR}.  As a local density-based method, DP can obtain good results in most instances. But as a centroid-based method, DP and its variants are unable to cluster
points correctly when a category has more than one centers.

The centroid-based methods focus on mining the centers and then assign the other points. This kind of method clusters data point by point. However, as intelligent humans, we prefer methods that can classify data cluster by cluster or part by part.

Schikuta \cite{SE} designed a grid in the data distribution area to partition data into blocks. The points in grid cells of greater density are considered to be members of the same cluster. Grid-based clustering had developed many extensions in recent years, such as grid ranking strategy based on
local density and priority-based anchor expansion \cite{DSQ}, density peaks clustering algorithm
based on grid \cite{XUX} and shifting grid clustering algorithm \cite{ME}. However, these grid-based methods cannot be applied on high-dimensional datasets as the number of cells in the grid grows exponentially with the dimensionality of data.

\section{Proposed methods}

In general, the cluster centers are surrounded by neighbors with lower local density, and a cluster center is at a
relatively large distance from other cluster centers. Base on this idea, we can make this algorithm assumption: there are $k-1$ neighbourhoods $U(x_i,r_i)$ composed of higher local density points in the dataset of $k$ categories. This assumption is satisfied in many existing simulation and real datasets.

Border points are distributed in two cases: ($i$) the border points of $i$th cluster are far away from the border points of $j$th cluster ($i\neq j$); ($ii$) the border points of different clusters are close together. We first apply  local density indicator for denoising, and then cluster objects for case ($ii$).

\subsection{Fission clustering algorithm}

In this section, we will deal with the case ($i$) first.

To develop the algorithm further, we present the following definition that
will be used throughout this article.

{\bf  Definition.} $f:X\times X\rightarrow R$ is a distance (similarity) function, where $X$ is a sample set, $R$ is the real number set. For all $x_k\in X$, if  $f(x_0,x_k)\not\in(f(x_0,x_i),f(x_0,x_j))$ (or $f(x_0,x_k)\not\in(f(x_0,x_j),f(x_0,x_i))$), we call $|f(x_0,x_i)-f(x_0,x_j)|$ a crack of $(X,f)$, where $x_0,x_i,x_j\in X$.

Obviously, the maximal crack ($MC$) of $(X,f)$  exists for a finite dataset.

The key steps of the FC algorithm are to fissure a dataset into two subsets and to stop fissuring subsets when all the clusters are obtained.  These two key steps are presented as follows.

\subsubsection{Dividing datasets}

Suppose $f(x_i,x_j)<f(x_i,x_k)$ if the relationship of $x_i$ and $x_j$ is closer than the relationship of $x_i$ and $x_k$.  The distance (similarity) matrix of $(X,f)$ can be obtained easily, and is denoted as $S(X)$. $S_1(X)$ is obtained by sorting every row of the distance matrix $S(X)$. The $i$th column of $S_1(X)$ is subtracted from the $(i+1)$th column to acquire the $i$th column of $S_2(X)$, $MC=max\{s_{ij}:s_{ij}\in S_2(X)\}$. Suppose $|f(x_i,x_j)-f(x_i,x_k)|=MC$, if $f(x_i,x_t)\leqslant min\{f(x_i,x_j),f(x_i,x_k)\}$, then $x_t\in X_1$; otherwise, $x_t\in X_2$, set $X$ is fissured into two subsets.

If there are $k$ categories of objects in $X$, the $k$ categories can be obtained step by step using the above fissuring method.

\subsubsection{Stop dividing datasets}

The fundamental and difficult task of clustering analysis is  determining the number of clusters. The number of categories is known as an assumption in the initial clustering research. A clustering approach with few input parameters is expected when we face increasing numbers of poorly information datasets  (scant or incomplete data).  Many studies have addressed this difficult issue in recent decades. In this paper, the characteristics of the distance matrix are investigated, and then the useful information in the matrix is applied to determine the number of categories.

We use the following formulae as illustration: let $d_0(C)=max\{f(x_i,\hat{x_i}):x_i\in C\subset X\}$ and $d_0=max\{f(x_i,\hat{x_i}):x_i\in X\}$, where $\hat{x_i}$ is the nearest neighbor of $x_i$. Suppose
there is a path such that $x_i$ and $x_j$ are connected for all $x_i,x_j\in C$, and the distance of every pair of connection points on the path is less than or equal to $d_0(C)$. This path is denoted as $d_0(C)$-path. The following Theorem is an effective indicator to determine the number of categories.\\

{\bf Theorem.} If the distance function $f$ satisfies triangle inequality and $C\subset X$ has a $d_0(C)$-path, then $MC(C)\leqslant d_0(C)$, where $MC(C)$ is the $MC$ of $(C,f)$.

{\textit{Proof.}} If there is a crack $|f(x_0,x_i)-f(x_0,x_j)|>d_0(C)$, then $f(x_i,x_j)\geqslant f(x_0,x_i)-f(x_0,x_j)>d_0(C)$ (suppose $f(x_0,x_i)>f(x_0,x_j)$). Thus, if $x_i$ and $x_j$ are not adjacent points on the $d_0(C)$-path, there must be a point $x_k$ on the path from $x_i$ to $x_j$ ($x_i\thicksim x_k\thicksim x_j$).

If $f(x_0,x_k)\not\in (f(x_0,x_j),f(x_0,x_i))$, then one of $|f(x_0,x_i)-f(x_0,x_k)|>|f(x_0,x_i)-f(x_0,x_j)|>d_0(C)$ and  $|f(x_0,x_j)-f(x_0,x_k)|>|f(x_0,x_i)-f(x_0,x_j)|>d_0(C)$ holds. Assuming that $|f(x_0,x_i)-f(x_0,x_k)|>d_0(C)$, then there is $x_t$ on the path $x_i$ to $x_j$ ($x_i\thicksim x_t\thicksim x_k\thicksim x_j$). Because the path of $x_i$ to $x_j$ is a part of the $d_0(C)$-path, $x_i$ and $x_j$ can be  connected by some points, and the distance between two connection points is less than or equal to $d_0(C)$. If $f(x_0,x_t)\not\in (f(x_0,x_j),f(x_0,x_i))$, then there must be a point $x_s$ on the path from $x_i$ to $x_j$ such that $f(x_0,x_s)\in (f(x_0,x_j),f(x_0,x_i))$ holds in the finite set $C$.

However, $|f(x_0,x_i)-f(x_0,x_j)|$ is a crack, $f(x_0,x)\not\in(f(x_0,x_j),f(x_0,x_i))$ for all $x\in C$. It is a contradiction. Hence, all the cracks must be less than or equal to $d_0(C)$.\\

If the distance of every pair of clusters is much greater than $d_0$, and every cluster has a $d_0$-path, the inequation $MC(C)\leqslant d_0$ can be considered as the condition under which stop fissuring a subset $C$. If all the subsets that fissured from $X$ are satisfied by the inequation $MC(C)\leqslant d_0$, the process of fissuring subsets will be stopped. The number of clusters will also be determined at the same time.

Numerous common distance functions satisfy triangle inequality, such as the Manhattan distance, Euclidean distance, and Minkowski distance.
If the densities of clusters are not extremely different in the same dataset, the inequation is effective.\\

The details of FC algorithm are shown as follows, where $S_k(C)(:,i)$ is the $i$th column of $S_k(C)$.

\rule{7.6cm}{0.5mm}

{\bf Algorithm 1:} FC algorithm.

\rule{7.6cm}{0.25mm}

{\bf Input:} Distance matrix $S(X)$.

{\bf Output:} Clusters of $X$.

1. $d_0\leftarrow max\{f(x_i,\hat{x_i}):x_i\in X\}$.

2. $C_1\leftarrow X$ (initial value).

3. {\bf While}  There is a subset $C_i$ such that $MC(C_i)>d_0$ {\bf do}

4.  {\bf repeat}

5.  Pick the subset $C_i$ if $MC(C_i)>d_0$.

6. Sort every row of $S(C_i)$ to obtain $S_1(C_i)$.

7. $S_2(C_i)(:,k)\leftarrow S_1(C_i)(:,k+1)-S_1(C_i)(:,k)$, $k=1,2,\cdots,n-1$.

8. $MC\leftarrow max\{S_2(C_i)(k,j):k=1,2,\cdots,n,\ j=1,2,\cdots,n-1\}$.

9. $|f(x_i,x_j)-f(x_i,x_k)|\leftarrow MC$.

10. If $f(x_i,x_t)\leqslant min\{f(x_i,x_j),f(x_i,x_k)\}$ then $x_t\in C_{count}$; otherwise, $x_t\in C_{count+1}$.

11.  {\bf until} $max\{MC(C_i)\}\leqslant d_0$.

12.  {\bf end while}

\rule{7.6cm}{0.25mm}\\

\subsection{The fission clustering algorithm with  k-nearest neighbors local density indicator (FC-KNN)}

% \subsubsection{Heat diffusion fission clustering algorithm}

The main idea of this section is to obtain a dense subset $C\subset X$ in the case ($ii$) that can make the distances between every pair of clusters in $C$ large enough and the distances between every pair of nearest neighbors small enough, and then apply Algorithm 1 to split the subset $C$.

\subsubsection{Obtaining the local density indicator}

The aim of this subsection is to obtain a local density indicator $\rho_i$ for every object $x_i$, and then distinguish the dense area objects from the sparse area objects.

A relatively straightforward method is utilized to obtain the local density indicator $\rho_i$, shown as follow equation,

\begin{equation}
\rho_i=1/\sum\limits_{x_j\in KNN(x_i)} f(x_i,x_j),
\end{equation}

where $KNN(x_i)$ is the k-nearest neighbor set of $x_i$.

To compare with the objects of sparse area, the objects in dense area have a spherical neighborhood with a smaller radius which contains the same number of neighbors. The object in dense areas is to obtain a larger local density indicator $\rho_i$ by using equation (1). The sample will be considered to belong to the dense subset $C$ if it has a larger $\rho_i$.

\subsubsection{The processes of FC-KNN algorithm}

The main steps of FC-KNN algorithm are shown as follows.

{\bf Step 1.} Use Algorithm 2 to obtain a dense subset $C$.

{\bf Step 2.} Cluster the subset $C$ by using Algorithm 1.

{\bf Step 3.} Assign the objects of $X-C$ to their nearest cluster.\\

A simple denoising method is shown as Algorithm 2.

\rule{8cm}{0.5mm}

{\bf Algorithm 2:} Denoising.

\rule{8cm}{0.25mm}

{\bf Input:} Distance matrix $S(X)$, parameters $t$ and $N_0=|KNN(x_i)|$.

{\bf Output:} The dense subset $C$.

{\bf Initialize:} r=0.4 (In general $|C|>[n/2]$, $n=|X|$).

1. Apply  equation (1)  to obtain $\rho_i$ for every object.

2. Remove $[0.4*n]$  objects of $X$ that have smaller $\rho_i$, retain the other objects in $C$.

3. $d_0\leftarrow max\{f(x_i,\hat{x_i}):x_i\in C\}$.

4. {\bf While} $MC(C)\leqslant t\times d_0$ {\bf do}

5. {\bf repeat}

6. $r\leftarrow r+0.1$.

7.  Remove $[r*n]$ objects of entire dataset $X$ that have smaller $\rho_i$,  retain the other points in $C$.

8.  Update $d_0$ and $MC(C)$ of the new subset $C$.

9.  {\bf until} $MC(C)>t\times d_0$ or $r=0.9$.

10. {\bf end while}.

\rule{8cm}{0.25mm}

When the fission of dense subset $C$ is complete after Step 2 of the FC-KNN processes, the remaining objects in set $X-C$ need to be assigned to their right category. A simple method is applied to assign the objects of $X-C$: let $A\subset X$ be the subset that contains the already classified points and  $U\subset X$ be the subset of unclassified points. If $f(x_i',x_j')=min\{f(x_i,x_j):x_i\in A, x_j\in U\}$, then $x_j'$ is assigned to the category that contains $x_i'$.

\begin{figure*}[!hbt]
\centering
\includegraphics[scale=0.56]{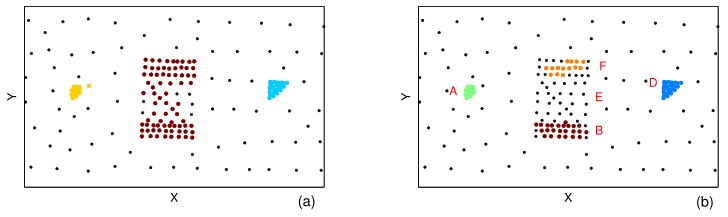}
\caption{Results for tuning course with parameter $t$.}
\label{}
\end{figure*}

Shown as FIGURE 1, $t>1$ can be considered as a tuning parameter. Algorithm 2 increases the value of $t$ to remove more border points (sparse area points). For $N_0=[2\%n]$, when $t\in(1,6]$ the dense subset $C=A\cup D \cup B \cup E \cup F$, the families $B$ and $F$ are connected by some points of $E$, so the dense families of categories are $A$, $D$ and $B \cup E \cup F$. When $t\in[7,11]$ the dense subset $C=A\cup D \cup B \cup F$, the points of $E$ are considered as the sparse area points and removed, so the dense families of categories are $A$, $D$, $B$ and $F$. When $t>12$ the dense subset $C=A\cup D$, the points of $B \cup E \cup F$ are considered as the sparse area points and removed, the dense families of categories are $A$ and $D$.

Equation (1) takes $\mathscr{O}(n*N_0)$ operations. Algorithm 1 splits the set $X$ (or dense subset $C$) into subsets $C_1,C_2,\cdots,C_k$. Since $|C_i|\ll |X|$, data processing will be faster and faster accompanied by the dividing courses of subsets. The dividing course only need to implement $k-1$ times to obtain $k$ clusters.

\section{Experiments}

In this section, we evaluate the performance of the proposed method on both simulation data and real data, and then compare it with some state-of-the-art methods that do not need  the number of clusters to be input. All the experiments are implemented based on the same software and hardware: MATLAB R2014a in Win7 operating system with Intel Core I5-3230M 2.6GHz and 12G Memory.

The Euclidean function was applied to obtain the distance matrix in all experiments. We selected the following methods for our comparisons with the proposed method: the affinity propagation algorithm (AP) \cite{FBJ}, fast search and find of density algorithm (DP) \cite{RA},  NK hybrid genetic algorithm (NKGA) \cite{TR} and Grid-clustered rough set (GCRS) model \cite{SML}.

\subsection{Descriptions of Experiment data}

\subsubsection{Simulation data}

\begin{figure*}[!hbt]
\centering
\includegraphics[scale=0.61]{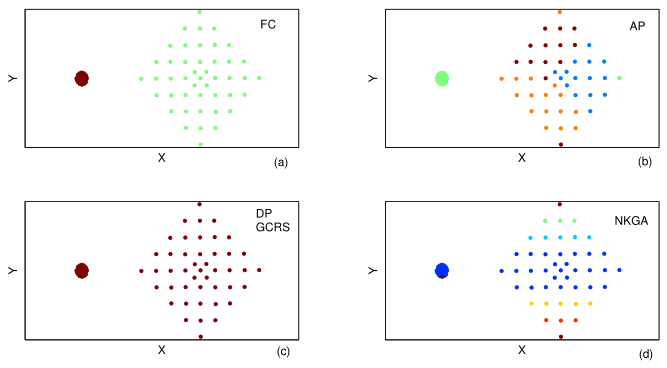}
\caption{The clustering results for different methods used on the Imbalance dataset. {\small ((a), (b) and (d) were clustered by FC, AP and NKGA, respectively. DP and GCRS  obtained the same result as in (c)).}}
\label{}
\end{figure*}

\begin{figure*}[!hbt]
\centering
\includegraphics[scale=0.6]{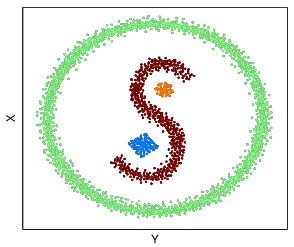}
\caption{Synthesis}
\label{}
\end{figure*}

First, some frequently-used datasets  obtained from different references are applied to test the algorithms, such as R15 \cite{VCJ}, A1 \cite{IK}, S1 \cite{FP}, Dim2 \cite{FP1} and Dimond \cite{SS} etc. And then two datasets, Imbalance (FIGURE 2) and Synthesis (FIGURE 3), are constructed for the supplementary tests. All the simulation data are points of two-dimensional Euclidean space.

\begin{table*}
\centering
\footnotesize
\label{tab:parametervalues}
\setlength{\tabcolsep}{3pt}
\begin{tabular}{|p{38pt}|p{23pt} p{25pt}|p{15pt}p{15pt}p{23pt}p{23pt} p{38pt}|p{25pt} p{26pt} p{28pt} p{32pt} p{38pt} |}
\hline
Dataset &  \multicolumn{2}{|c|}{Details}  & \multicolumn{5}{|c|}{Number of clusters}   & \multicolumn{5}{|c|}{Accuracy} \\
& objects & clusters  & AP & DP & NKGA & GCRS& FC-KNN &  AP & DP  &NKGA & GCRS&FC-KNN  \\
\hline
D31 &   3100  &31 & 8  & {\bf 31 }& 19 & {\bf 31} & {\bf 31}  & 0.2045  &  0.3474 &   0.3442  &  0.9335  &  {\bf 0.9677} \\
 Flame &  240  & 2 & 3 &{\bf  2 }& {\bf 2} &{\bf 2 }& {\bf 2} &  0.7167  &  0.3000  &  0.6583 &   0.8625  &  {\bf 0.9958}\\
R15 &  600 &15  &5 & {\bf 15}&{\bf 15}&{\bf 15} &{\bf 15} & 0.1867  &  0.9800  &  0.8983 &   0.9800 &  {\bf  0.9933}\\
 Dimond &  2999  &9  &15 &{\bf  9}&5&{\bf 9 }&{\bf 9 } & 0.3178   & 0.6889 &   0.5552  &  {\bf 1.0000}  & {\bf  1.0000}\\
 Imbalance &   101  &2 & 4&1 &6 &1&{\bf 2} & 0.6931  &  0.5545  &  0.5941 &   0.5545  & {\bf  1.0000}\\
Synthesis &  2461  &4& 16 &10 & 21& 3 & {\bf 4}  & 0.2751 &   0.2670  &  0.3946  &  0.5095  & {\bf  1.0000}\\
\hline
\end{tabular}
\caption{The description of data and the results comparison of different methods}
\label{tab1}
\end{table*}

\begin{center}
\footnotesize
\begin{table*}[htb]
%\label{tab:parametervalues}
%\setlength{\tabcolsep}{3pt}
\begin{tabular}{|p{38pt}|p{32pt} p{32pt} p{32pt}|p{306pt}|}
\hline
Dataset& Instances &Features & Clusters  & Detail  \\
\hline
Iris &   150 & 4 &3 &50 Iris Setosa, 50 Iris Versicolour and 50 Iris Virginica \\
\hline
Seeds &  210 & 7 & 3 & seeds from Kama,  Rosa and Canadian, 70 seeds each place \\
\hline
Vertebral &  310 &6  &2   & 210 abnormal and 100 normal   \\
\hline
Wifi &  2000 &7 &4  & 2000 times of signals records in 4 rooms, 500 records each room  \\
\hline
Adenoma &   6 &12488 &2 & 6 genes: 3 ADE and 3 N1\\
\hline
Myeloid & 21 & 22283 & 3& 21 genes: 12 acute promyelocytic leukemia (APL) genes, 3 polyploidy genes of APL and 6 Acute myeloid leukemia genes\\
\hline
\end{tabular}
\caption{The simple description of real data}
\label{tab1}
\end{table*}
\end{center}
\normalsize

\subsubsection{Real data}

Several real-world datasets are applied to test the performance of the proposed method, two datasets for plant shape recognition: Iris\footnote{http://archive.ics.uci.edu/ml/datasets.php} \cite{FRA, HF} and Seeds$^1$ \cite{CM}; a wireless signal dataset: Wifi$^1$ \cite{RJG}; a human vertebral column dataset: Vertebral$^1$ \cite{BEE}; and two gene datasets: Adenoma\footnote{http://portals.broadinstitute.org/cgi-bin/cancer/datasets.cgi} \cite{SCA} and Myeloid$^2$ \cite{STK}. Two gene datsets were taken from Cancer Program Datasets and others were taken from the UCI repository. Simple descriptions of these real datasets are provided in Table 2.

\subsection{Comparisons and Discussions}

\begin{table*}[htb]
\centering
\footnotesize
\label{tab:parametervalues}
\setlength{\tabcolsep}{3pt}
\begin{tabular}{|p{32pt}|p{15pt}p{20pt}p{22pt}p{20pt} p{38pt}|p{22pt}p{28pt}p{22pt}p{22pt} p{38pt}|p{22pt}p{28pt}p{22pt}p{22pt} p{38pt}|}
\hline
Dataset &  \multicolumn{5}{|c|}{Number of clusters}   & \multicolumn{5}{|c|}{Accuracy} & \multicolumn{5}{|c|}{F-Score}\\
&  AP & DP & NKGA &GCRS &FC-KNN  &  AP & DP  &NKGA & GCRS &FC-KNN &   AP & DP  &NKGA& GCRS &FC-KNN  \\
\hline
Iris &   2  & 2 & 11 & {\bf 3} & {\bf 3}  &      0.5333 &   0.6667  &  0.4533  & {\bf  0.9467}& 0.9067  &  0.4329  &  0.5714  &  0.5883 &  {\bf  0.9503} &   0.9168 \\
 Seeds &  2 & {\bf 3} & 2 &{\bf 3} & {\bf 3} &   0.6048  &  0.6857 &   0.3286 &   0.8524  &  {\bf 0.8857} &   0.5102  &  0.8007  &  0.1649 &   0.8575  &  {\bf 0.8900}\\
Vertebral & 1 & 1&{\bf 2}&{\bf 2} &{\bf 2} &   0.6774  &  0.6774  &  0.6645  &  0.5258  &  {\bf 0.7710}  &  0.4038 &   0.4038  &  0.3992  &  0.6752 &   {\bf 0.7976}\\
 Wifi &  5 & {\bf 4}&1&3 &{\bf 4}  &  0.1405  &  0.1025  &  0.2500  &  0.7450  &  {\bf 0.9355} &   0.1671  &  0.1859  &  0.1000 &   0.6777   & {\bf 0.9402}\\
Adenoma & 1 &{\bf 2} &{\bf 2} & 4 &{\bf 2 }&   0.5000  & {\bf 1.0000 } &  0.6667 &   0.6667  & {\bf 1.0000 }&  0.3333 &  {\bf 1.0000 } &  0.7273  &  0.8000   & {\bf  1.0000}\\
Myeloid &  1 &2 &{\bf 3} &1 & 2  &  0.5714  &  0.6190  & {\bf 0.7143 }&   0.5714 &  {\bf 0.7143}& 0.2424  &  0.4896   & {\bf 0.7430}  &  0.2424  &  0.6061\\
\hline
\end{tabular}
\caption{The results comparison of different methods}
\label{tab1}
\end{table*}

The clustering results for the simulation data and real data are shown in Table 1 and Table 3, respectively.

In FIGURE 2 and 3, no single point can be considered as the geometrical centroid of the annulus in the Synthesis dataset, the densities of the two clusters in the Imbalance dataset  have a large difference. It is difficult to determine the number of categories with these AP, DP, NKGA and GCRS algorithms. DP and GCRS can not find the second center point of Imbalance dataset, then these two algorithms consider it as one cluster dataset.  The proposed method is aimed at mining the dense family of every category, not the center points, so the proposed method can correctly determine the number of clusters for the Imbalance and Synthesis datasets.

To evaluate and compare the performance of the clustering methods, we apply the evaluation metrics: Accuracy  and F-score \cite{DUL}
in our experiments to do a comprehensive evaluation.  The higher the value, the better the clustering performance for the two measures.
Comparing with the best results of other algorithms  indicates that our method has  relative advantages of 0.19 and 0.2625 (TABLE 3) with respect to Accuracy and F-Score for Wifi dataset, respectively.

In the description of algorithms, the distance matrix is a significant input. The distance matrix depends on the correct selection of attributes, correct value of selected attributes and a good distance (recognition) function. The recognition function $f_1$ is  better (stronger) than $f_2$ if $|f_1(x_i,x_i')-f_1(x_i,x_j)|\geqslant|f_2(x_i,x_i')-f_2(x_i,x_j)|$ for all $x_i,x_i'\in C_i$ and $x_j\in C_j$, where $C_i$ and $C_j$ are two clusters of $X$.

The simulation data are Euclidean space points, the Euclidean function is a strong recognition function for them, then the parameter $t$ can be set to a large value. If the Euclidean function is a weak recognition function for some real datasets, such as the dataset Vertebral, the AP algorithm classifies Vertebral as one cluster.  FC-KNN can determine the right clusters after tuning parameter $t$ with a smaller value when the recognition function is week.

The dataset Adenoma is a great challenge in clustering analysis, with especially small sample size and extremely high sample dimensionality. It is very difficult to determine the center point of every cluster, but it is easy to distinguish between the points with a small value of $\rho_i$, hence, FC-KNN can obtain the correct clusters.

The proposed method is robust, it can obtain a same clustering result when we select values for parameters $t$ and $N_0$ with wide intervals $[t^-,t^+]$ and $[N_0^-,N_0^+]$,  respectively. When we face poor information datasets, the methods needed to input the number of clusters are infeasible, our method still works. The parameters of our algorithm are easy to set, $N_0$ is established by the indicator of sample number, and $t$ can be tuned to obtain the results needed.
The stronger the distance (recognition) function the easier the selection of parameters. All  simulation data obtained from different references use the value $N_0=ceil(2\%n)$ or $N_0=ceil(3\%n)$ and $t=4$.

In a word, our method  obtain a better results with respect to the estimation of cluster number, Accuracy and F-Score, compared with other methods.

\section{Conclusion}

The data clustering  courses of many current methods are similar to the courses of atomic fusion, we have proposed a method for data clustering based on atomic fission patterns. Different from existing clustering methods which focus on seeking one center point of every category, the proposed algorithm acquires dense families  of categories.   The idea of our method is to apply spherical neighborhood  instead of  grid cells to cover the distribution space of objects, and the method  needs to determine only $k-1$ spherical neighborhoods for a $k$ categories dataset. Hence, it will not be influenced by the data dimension, unlike  grid-based clustering. Experimental results on simulation data and real data reveal the  effectiveness of the proposed method. In future research, we aim to extend the proposed algorithm in order to cluster more kinds of datasets that overstep the scope of preamble assumptions.



%\end{multicols}

\end{multicols}{2}
\end{document}